\documentclass[runningheads]{llncs}

\usepackage{accv}

\usepackage{accvabbrv}

\usepackage{graphicx}
\usepackage{booktabs}

\usepackage[accsupp]{axessibility}  

\usepackage{tabularx}
\usepackage{amsmath}
\usepackage{multirow}
\usepackage{array}

\usepackage{hyperref}

\usepackage{orcidlink}

\definecolor{light-gray}{gray}{0.91}
\newcommand{\mathcolorbox}[2]{\colorbox{#1}{$\displaystyle #2$}}

\newcommand{\parlabel}[1]{\vspace{0pt}\noindent\textbf{#1}}

\usepackage{tabularx}

\begin{document}

\title{Designing Extremely Memory-Efficient CNNs for On-device Vision Tasks}

\titlerunning{Memory-Efficient CNNs for On-device Vision}
\author{Jaewook Lee\orcidlink{0009-0005-1703-7971} \and
Yoel Park\orcidlink{0009-0000-1600-934X} \and
Seulki Lee\orcidlink{0009-0004-7162-0845}}

\authorrunning{Jaewook Lee, Yoel Park, and Seulki Lee}

\institute{Ulsan National Institue of Science and Technology (UNIST) \\
\email{jaewook.lee@unist.ac.kr, joel157@unist.ac.kr, seulki.lee@unist.ac.kr}}

\maketitle
\begin{abstract}
In this paper, we introduce a memory-efficient CNN (convolutional neural network), which enables resource-constrained low-end embedded and IoT devices to perform on-device vision tasks, such as image classification and object detection, using extremely low memory, \ie, only 63 KB on ImageNet classification. Based on the bottleneck block of MobileNet, we propose three design principles that significantly curtail the peak memory usage of a CNN so that it can fit the limited KB memory of the low-end device. First, `input segmentation' divides an input image into a set of patches, including the central patch overlapped with the others, reducing the size (and memory requirement) of a large input image. Second, `patch tunneling' builds independent tunnel-like paths consisting of multiple bottleneck blocks per patch, penetrating through the entire model from an input patch to the last layer of the network, maintaining lightweight memory usage throughout the whole network. Lastly, `bottleneck reordering' rearranges the execution order of convolution operations inside the bottleneck block such that the memory usage remains constant regardless of the size of the convolution output channels. The experiment result shows that the proposed network classifies ImageNet with extremely low memory (\ie, 63 KB) while achieving competitive top-1 accuracy (\ie, 61.58\%). To the best of our knowledge, the memory usage of the proposed network is far smaller than state-of-the-art memory-efficient networks, \ie, up to 89x and 3.1x smaller than MobileNet (\ie, 5.6 MB) and MCUNet (\ie, 196 KB), respectively.
\end{abstract}

\section{Introduction}

\begin{figure} [tb]
  \centering
  \includegraphics[width=0.7\textwidth]{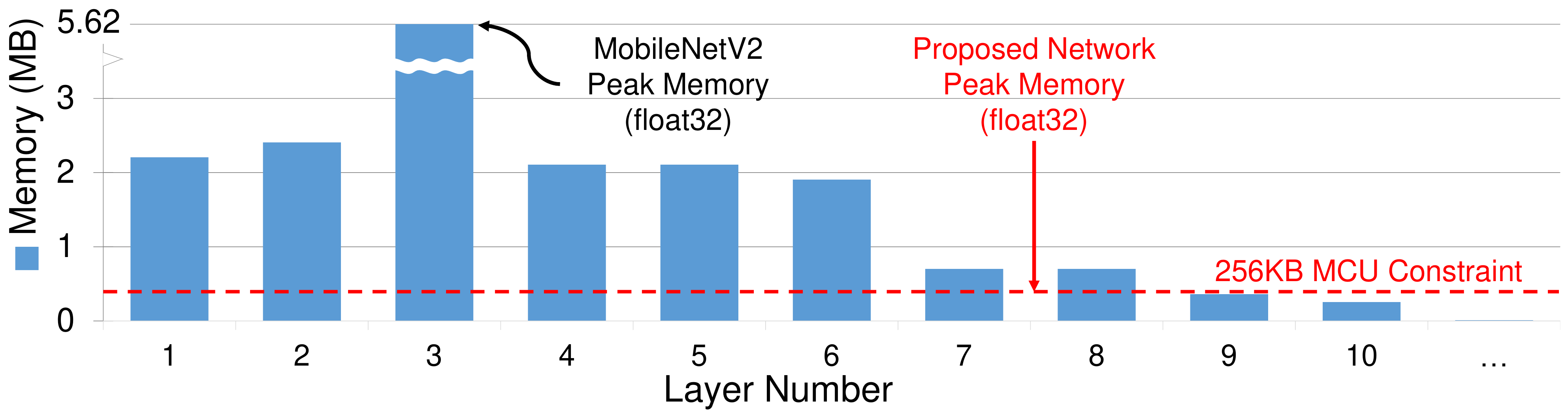}

  \caption{Peak memory usage across MobileNetV2 layers on ImageNet 224x224, reaching 5.62MB at the third. The red line indicates our network's 256KB peak memory in fp32.}
  \label{fig:memory}
\end{figure}

As an increased number of vision applications are moving towards low-end devices, a variety of on-device CNNs have been devised in an efficient and lightweight manner. Among many of them, MobileNet~\cite{sandler2018mobilenetv2} has become the de facto standard for many resource-constrained embedded, mobile, and IoT devices thanks to its efficient network architecture, especially the simple yet effective bottleneck block. Although MobileNet~\cite{sandler2018mobilenetv2} enables low-end devices to run various vision tasks(\eg, image classification~\cite{deng2009imagenet} and object detection~\cite{Everingham2015,lin2015microsoft}), the stacked bottleneck block structure significantly contributes to its peak memory (RAM) requirement(\ie, 5.6 MB for ImageNet~\cite{deng2009imagenet}), which is still far exceeds the limited memory capacity of millions of low-end devices having only several hundreds of KB of RAM(\eg, 256 KB), as shown in \cref{fig:memory}. Considering that 1) embedded cameras are becoming cheaper, smaller, more powerful, and low-powered~\cite{choi2015always,bigas2006review}, and 2) an explosive number of embedded cameras are being installed on low-end devices at a fast pace, an increasing number of low-end devices and cameras are expected to handle advanced vision tasks on the device.

The bottleneck block of MobileNet is the key enabler for the rapid deployment of CNNs onto resource-constrained low-end devices. Based on an inverted residual structure where the skip connections are between the bottleneck layers, it efficiently learns features and representations by applying point-wise and depth-wise convolutions. However, since it deals with large-sized images with the bottleneck block that expands the intermediate features with the point-wise (expansion) convolution, its memory usage soars (5.6 MB) beyond the memory capacity of many low-end devices having only KB of memory, such as ARM core-based IoT devices~\cite{STM32F,STM32H7}. Therefore, to meet the growing demand for on-device vision applications running on low-end devices, the memory usage of MobileNet, composed of multiple bottleneck blocks, should be reduced substantially.

In this paper, we introduce an extremely memory-efficient CNN, which dramatically reduces the memory usage of the bottleneck block~\cite{sandler2018mobilenetv2}, making MobileNet's learning capability accessible to memory-constrained low-end devices for various computer vision tasks(\eg, image classification~\cite{deng2009imagenet,chowdhery2019visual} and object detection~\cite{Everingham2015,lin2015microsoft}). We propose three design principles that collectively construct the proposed memory-efficient(and memory-aware) CNN, with peak memory usage can be flexibly adjusted to fit the limited memory budget of low-end devices. \textbf{First}, `input segmentation' divides an input image into a set of patches, reducing the memory requirement of large input images at the beginning of the network. To compensate for the possible performance degradation caused by input segmentation, we make patches partially overlap with each other along their edges and add a central patch to capture crucial features in the center of the image. \textbf{Next}, each segmented input patch constructs a dedicated network path, consisting of multiple bottleneck blocks, all the way through the whole network independent of each other, which we call `patch tunneling', keeping the memory usage of each patch under the memory budget. At the last layer of the network, they are combined to cooperate for the target task(\eg, image classification and object detection). \textbf{Lastly}, `bottleneck reordering' maintains the memory usage of each bottleneck to be constant by rearranging execution orders of the point-wise and depth-wise convolution operations inside the bottleneck. Unlike existing approaches that execute convolution operations layer by layer without managing the peak memory usage, it reorders them across layers in a memory-efficient manner based on the computational properties of the convolution.

We implement the proposed network and deploy it onto a real low-end device, \ie, STM32H7~\cite{STM32H7}. We conduct experiments with two classes of vision tasks, \ie, image classification (ImageNet~\cite{deng2009imagenet} and VWW~\cite{chowdhery2019visual}) and object detection (PASCAL-VOC~\cite{Everingham2015} and MS-COCO~\cite{lin2015microsoft}). The results show that the proposed network enables on-device vision tasks with extremely low memory while providing competitive model performance, \eg, 61.58\% and 63.84\% top-1 accuracy on ImageNet using only 63 KB and 254 KB of memory, respectively, validating the efficacy of the proposed network for memory-constrained low-end devices.

The contributions of this paper are summarized as follows.
\begin{itemize}
    \item{We introduce an extremely memory-efficient CNN for memory-limited low-end devices, taking only 63 KB of memory for ImageNet classification, which is 89x and 3.1x lower than state-of-the-art memory-efficient CNNs, \ie, MobileNet (5.6 MB)~\cite{sandler2018mobilenetv2} and MCUNet (196 KB)~\cite{lin2021mcunetv2}, respectively.}
    \item{We propose three memory-aware design principles, \ie, `input segmentation', `patch tunneling', and `bottleneck reordering', which construct an extremely memory-efficient CNN based on the memory budget of low-end devices.}
    \item{We implement and deploy the proposed network onto a real embedded device (\ie, STM32H7~\cite{STM32H7}), demonstrating that it enables extremely memory-efficient on-device vision tasks, \ie, image classification and object detection, achieving 61.58\% top-1 accuracy on ImageNet using only 63 KB memory.}
\end{itemize}

\section{Related Work}
\parlabel{General Techniques for Making Networks Efficient}: Various methods have been proposed to improve the efficiency of neural networks. Quantization~\cite{gholami2022survey,liang2021pruning} simplifies the precision of number formats used in a network, usually from float32 to int8, int4, int2, or even binary bit~\cite{huang2019efficient}, intending to reduce the amount of computation. Although it can decrease memory usage along with computation, the memory reduction ratio is bounded to the number of bits quantized, usually 4x (from float32 to int8), and the model performance tends to deteriorate as more aggressive quantization is applied~\cite{lin2024awq}. Pruning~\cite{xu2020convolutional,vadera2022methods} is another technique that removes unnecessary weight parameters in a network based on their importance~\cite{han2015deep}. After pruning, a network becomes streamlined, improving resource efficiency, including the memory requirement. However, it deforms the original network architecture, causing several negative impacts, such as model performance degradation, unexpected model behavior, and repeated training procedures. Even worse, the memory usage may not be reduced after pruning if the pruned weight parameters are replaced with zeros, leaving the memory usage for the related operations the same. NAS (Neural Architecture Search)~\cite{dhar2021survey,he2021automl,elsken2019neural} seeks the optimal network architecture, which can be applied to find a memory-efficient network by imposing a memory constraint during the search~\cite{liberis2021munas}. Although it automatically searches for a potentially memory-efficient network, it is not guaranteed to find the best architecture that satisfies both the model performance and memory constraint. Moreover, the search process usually takes tens or hundreds of GPU days~\cite{zoph2018learning}, which is exorbitant in practice. Unlike those techniques orthogonal to the proposed methods, the three design principles introduced in this paper 1) construct a CNN with flexible memory reduction, not restricted by some upper bound, unlike quantization, 2) do not severely degrade the model performance, 3) readily apply to various tasks without modifying the network architecture, and 4) do not entail a long search time nor repeated training processes as NAS. Since they are orthogonal to the proposed methods, we apply one of them, \ie, quantization, to our CNNs.

\parlabel{MCUNet}~\cite{lin2020mcunet,lin2021mcunetv2} is a memory-efficient CNN proposed to perform ImageNet~\cite{deng2009imagenet} classification under 242 KB of memory on the STM32F board~\cite{STM32F}, achieving 60.3\% and 64.9\% top-1 classification accuracy in their first and second implementation, respectively. The memory requirement is reduced with the in-place depth-wise convolution and patch-based inference, executed by a run-time library called TinyEngine on the device. The network architecture that fits the target memory is searched by NAS~\cite{dhar2021survey,he2021automl,elsken2019neural} via the joint neural architecture and inference scheduling search. Although MCUNet is the first network that classifies ImageNet with the tight memory budget, it relies on existing techniques such as NAS and quantization, along with a significant embedded systems engineering procedure, entailing a large amount of time, computation, and human endeavors to find the network that fits the target memory budget. However, unlike MCUNet, the proposed network does not require quantization as mandatory and still achieves a similar level of memory usage, \ie, 254 KB with the float32 format, which can be further reduced to 63 KB when applying int8 quantization. Also, given a memory budget, the proposed network can be easily and flexibly constructed for various vision tasks, \eg, image classification and object detection, by applying the proposed three design principles accordingly, without requiring significant time, computation, and engineering efforts.
\section{Method: Design Principles}
\label{sec:method}
\begin{figure}[!t]
  \centering
  \begin{minipage}[t]{0.49\columnwidth}
    \centering
    \includegraphics[width=\columnwidth]{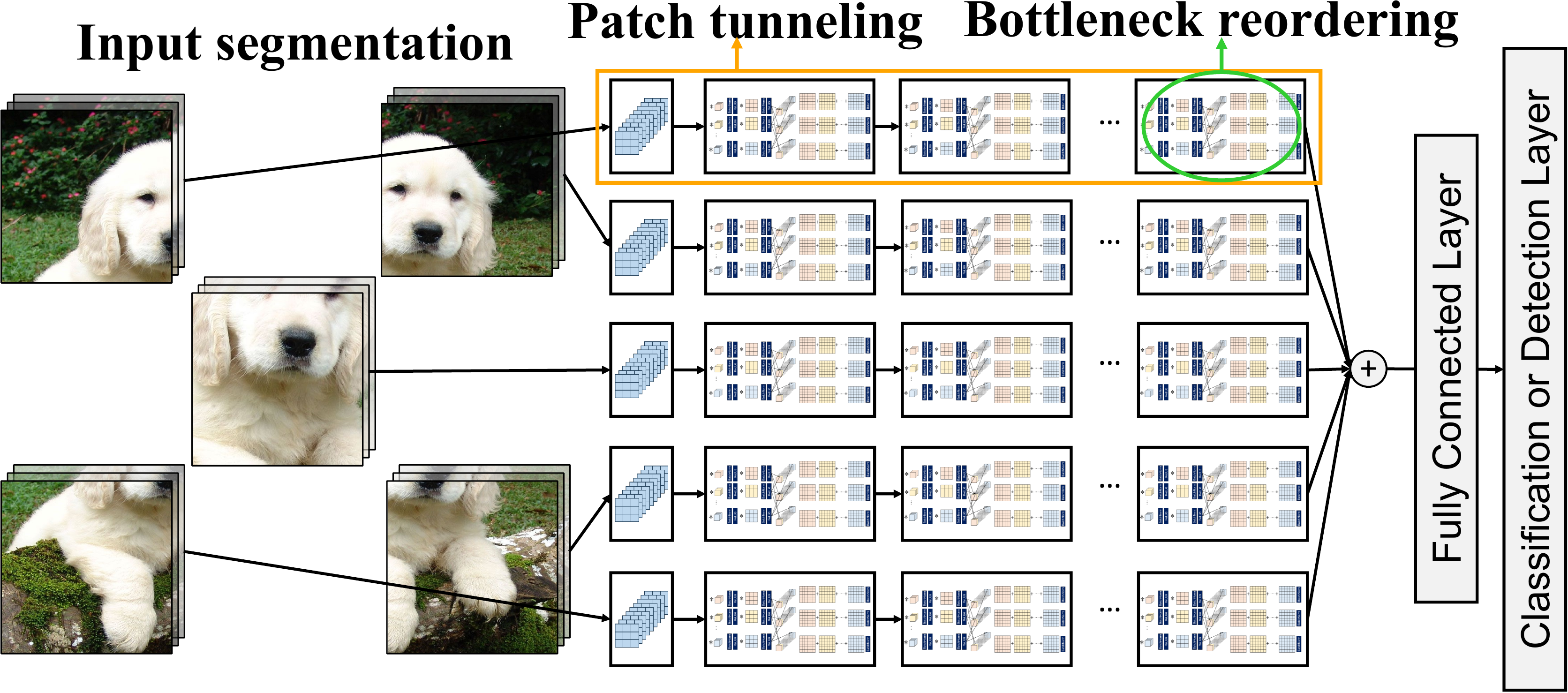}
    \caption{An overview of the proposed memory-efficient CNN constructed by the three memory-aware design principles: `input segmentation', `patch tunneling', and `bottleneck reordering'.}
    \label{fig:overview}
  \end{minipage}%
  \hfill
  \begin{minipage}[t]{0.49\columnwidth}
    \centering
    \includegraphics[width=\columnwidth]{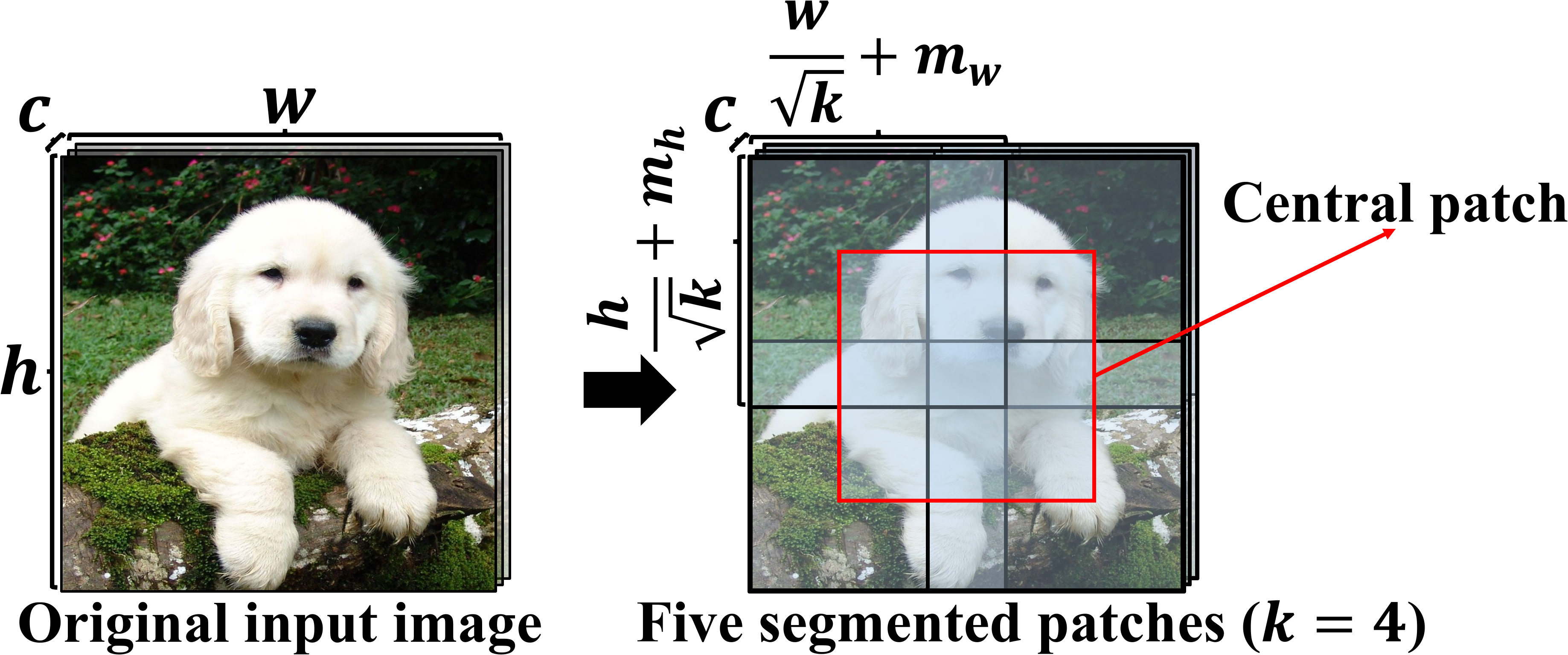}
    \caption{The proposed `input segmentation' splits the input image into $k$ patches, along with the central patch. Which results in a reduction of the initial memory requirement of the network into $\frac{1}{k}$.}
    \label{fig:input_segmentation}
  \end{minipage}
\end{figure}

\cref{fig:overview} depicts an overview of the proposed network, in which the peak memory usage of convolutional operations is flexibly adjusted under the memory budget of the target low-end device. It is achieved by the three memory-aware design principles, \ie, `input segmentation', `patch tunneling', and `bottleneck reordering', of which details are described in the following subsections.
\subsection{Input Segmentation}
As the size of images fed into a CNN is usually too large to fit the limited memory of many low-end devices, we decrease the dimension of the raw input image by dividing it into several patches through a process called `input segmentation' (\cref{fig:input_segmentation}). Given an input image of dimension $[ h \times w \times c]$, where $h$ is the height, $w$ is the width, and $c$ is the number of channels, the input image is segmented into $k$ patches of dimension $[ \frac{h}{\sqrt{k}} \times \frac{w}{\sqrt{k}} \times c ]$. For instance, the images in ImageNet~\cite{deng2009imagenet} (\ie, $[ 224 \times 224 \times 3 ]$), are segmented into four patches of the dimension $[ 112 \times 112 \times 3 ]$ when $k=4$, reducing the memory space required for the raw image by $\frac{1}{k}=\frac{1}{4}$ by processing each patch separately. 
Although such a simple segmentation can mechanically reduce the memory usage to $\frac{1}{k}$, it may degrade the model performance due to the segregation of features, making it difficult for the segmented patches to have a common view of the entire image. To compensate for such potential performance loss likely to be caused by segmentation, we add some margins along patch edges so that they can overlap slightly with each other. With the vertical and horizontal margins, denoted as $m_h$ and $m_w$, respectively, the dimension of patches then becomes:
\begin{equation}
    [ ( {h}/{\sqrt{k}} + m_{h} ) \times ( {w}/{\sqrt{k}} + m_{w} ) \times c ]
    \label{eq:margin}
\end{equation}
which makes the overlapped areas of $2m_{h}$ and $2m_{w}$ between two neighboring patches over the vertical and horizontal axis, respectively. For simplicity, by assuming that an image is square, \ie, $h=w$, and $m_{h} = m_{w}$, the ratio between the original image $[ h \times w \times c ]$ and the patch in \cref{eq:margin} becomes $\frac{1}{k} + \frac{2 m_{h}}{h \sqrt{k}} + \frac{m_{h}^{2}}{h^{2}}$, 
which becomes smaller when $k$ and $h$ increase, converging to $\frac{1}{k}$ when $h \gg m_{h}$.
For further improvement, we add the $(k+1)$-th patch on top of $k$ patches, which we call the central patch, to learn critical information located at the central part of the image likely to affect the model performance significantly, resulting in a total of $k+1$ patches. It is based on the observation that objects tend to be situated at the central area of the image so it is worth having a dedicated patch for that region. The central patch overlaps entirely with other $k$ patches, learning important features in the central area at the cost of having an additional convolution kernel for it. However, since each patch is executed independently, adding the central patch can be a useful way to enhance the model performance without the peak-memory increase as shown in the experiment (\cref{sec:experiment}), which can be regarded as a good trade off between the huge model performance improvement and the relatively small increase in the network size.
\subsection{Patch Tunneling}
The idea of `patch tunneling' is that features presented at each $k+1$ individual patches divided by `input segmentation' can be learned separately all the way through isolated layers of lightweight (memory-efficient) convolutions, which we call a patch tunnel, and then be effectively combined at the last layer of the network. It is different from conventional approaches that apply the convolution operation to the non-segmented entire image with a large number of convolution kernels (filters), which requires a huge memory space for storing the input and output of the convolution, instead of learning a smaller patch with a relatively small number of kernels taking a small amount of memory. With $k+1$ patches divided by `input segmentation' at the first layer of the network, `patch tunneling' constructs $k+1$ independent patch tunnels consisting of multiple layers of lightweight convolutions in the form of bottleneck~\cite{sandler2018mobilenetv2}, starting from each patch, such that the memory requirement of the convolution operation of each layer does not exceed the memory budget $m_{b}$ until the last layer of the network.
Given the input feature of dimension $[ h \times w \times c ]$ and the convolutional kernel (filter) as $[ k_{h} \times k_{w} \times c_{in} \times c_{out} ]$  at an arbitrary layer of a patch tunnel, the output dimension of the convolutional operation becomes $[ \bigl( \frac{h - k_{h} + 2p_{h}}{s} + 1 \bigl) \times \bigl( \frac{w - k_{w} + 2p_{w}}{s} + 1 \bigl) \times c_{out} ]$, where $s$ is the stride, and $p_{h}$ and $p_{w}$ is the vertical and horizontal padding, respectively. Since both the input and output should be stored in memory during the convolution operation, the summation of the input and output dimension, while ignoring the minute memory usage required for the convolution computation, should be maintained at most the target memory budget $m_{b}$ as:
\begin{equation}
    \underbrace{h \times w \times c}_{\text{Input}} + \underbrace{\bigl( \frac{h - k_{h} + 2p_{h}}{s} + 1 \bigl) \times \bigl( \frac{w - k_{w} + 2p_{w}}{s} + 1 \bigl) \times c_{out}}_{\text{Output}} \leq m_{b}
    \label{eq:input_output}
\end{equation}
which first starts from the segmented patches in \cref{eq:margin} as the initial input $[h \times w \times c]$ at the first layer of the network and computes the corresponding convolution output that will be used as the input for the next convolution layer.
By adjusting $s, k_{h}$, $w_{h}$, $p_{h}$, $p_{w}$, and $c_{out}$ accordingly, \cref{eq:input_output} becomes satisfied with $m_{b}$ for all convolution layers of each patch tunnel. Since the dimension of each patch is approximately reduced to $\frac{1}{k}$ of the original image with some margins as in \cref{eq:margin}, it is possible to find $s$, $k_{h}$, $w_{h}$, $p_{h}$, $p_{w}$, and $c_{out}$ satisfying \cref{eq:input_output} without degrading the model performance substantially. For example, the summation of the input and output in \cref{eq:input_output} becomes under $m_{b}$ = 256 KB (the size of SRAM on STM32F~\cite{STM32F}), by setting $s=1$, $k_{h}=3$, $w_{h}=3$, $p_{h}=0$, $p_{w}=0$, and $c_{out}=3$, given the input patch of ImageNet~\cite{deng2009imagenet} as $[ \bigl( \frac{h}{\sqrt{k}} + m_{h} \bigl) \times \bigl( \frac{w}{\sqrt{k}} + m_{w} \bigl) \times c ]$ with $h=224, w=224, c=3, k=4$, and $m_{h}=m_{w}=18$.

At the last layer of the network, the features independently learned and propagated through each patch tunnel are combined as the aggregated feature set with a fully connected layer. We found that summation is sufficient to aggregate features effectively, achieving competitive model performance while taking smaller memory compared to alternative aggregation methods such as concatenation. By keeping the memory requirement of each patch tunnel at most the memory budget $m_{b}$ and postponing the aggregation process of features learned in each tunnel until the last layer of the network, a single large convolution consuming a large amount of memory, can be effectively split into $k+1$ smaller patch tunnels, keeping the overall memory usage by almost $\frac{1}{k}$ in the entire network. 

\begin{figure}[!b]
  \centering
  \includegraphics[width = 0.8\columnwidth]{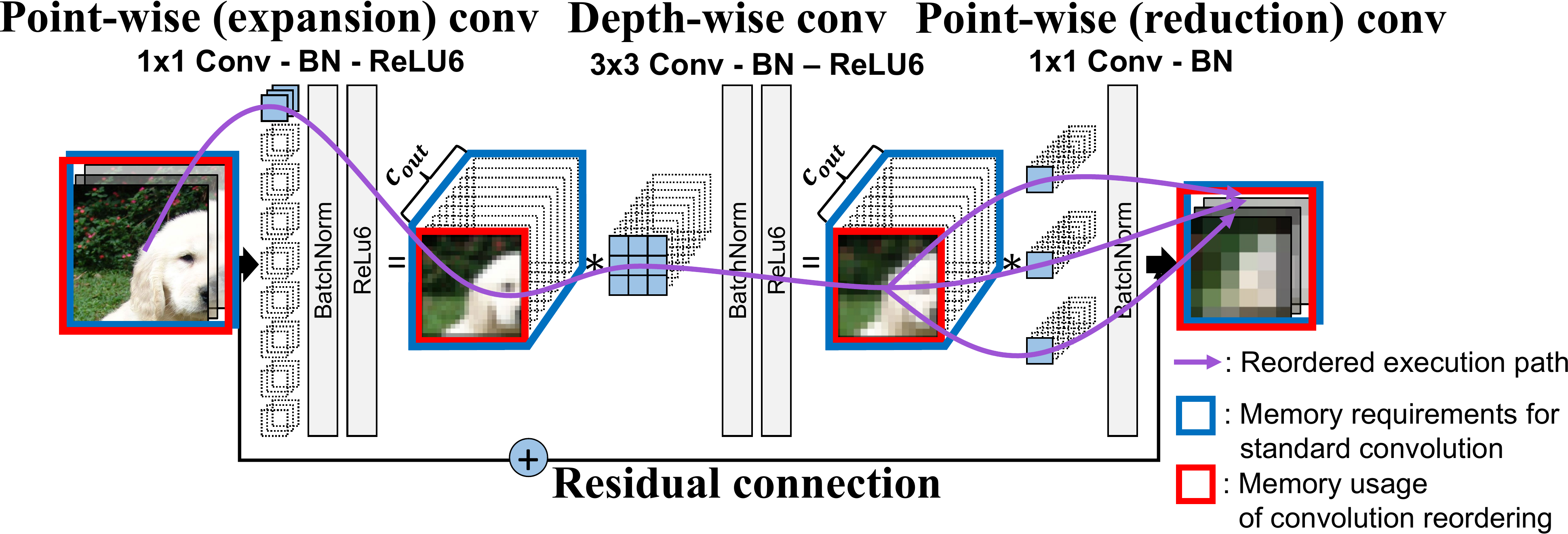}
  \caption{The proposed `bottleneck reordering' restricts the memory usage of the bottleneck to be constant irrespective of the output channel size ($c_{out}$) of the point-wise (expansion) and depth-wise convolution by rearranging their execution order: each convolution output channel is computed one at a time (red box), not all at once (blue box).
  }
    \label{fig:bottleneck_reordering}
\end{figure}
\subsection{Bottleneck Reordering}
Although each patch tunnel can retain its memory usage within the memory budget \(m_{b}\) by adjusting convolution kernel parameters, \ie, $s, k_{h}$, $w_{h}$, $p_{h}$, $p_{w}$, and $c_{out}$ in \cref{eq:input_output}, the extremely low memory budget of many low-end devices (e.g., \(m_{b}=256\) KB) limits the number of output channels \ie, $c_{out}$, to a point where feature learning becomes insufficient, potentially degrading model performance. In \cite{sandler2018mobilenetv2}, the expansion ratio \(t\) was introduced to expand \(c_{out}\) in point-wise (expansion) convolution. Then $c_{out}$ can be expressed in terms of $t$, such as $c_{out} = c_{in} \times t$. For adequate model performance, however, \(c_{out}\) must be increased beyond the constraints of the memory budget \(m_{b}\).

To tackle this problem, we propose `bottleneck reordering', which rearranges the operation order of the convolution in the bottleneck~\cite{sandler2018mobilenetv2} such that \cref{eq:input_output} is satisfied with a large $c_{out}$. It executes the convolution operation for each output channel independently one at a time over the three convolution layers in the bottleneck, \ie, the point-wise (expansion), depth-wise, and point-wise (reduction) convolution, until all the output channels are computed, as shown in \cref{fig:bottleneck_reordering}. Then, the independently computed outputs of each channel are relayed to the last layer of the bottleneck and accumulated into the final output. Since only a single output channel is computed and stored in memory one at a time over the entire bottleneck block, the memory space required to store the outputs of the point-wise (expansion) and depth-wise convolution in \cref{eq:input_output} is reduced to $\frac{1}{c_{out}}$ for each. Unlike the conventional way that performs convolutions over all output channels at the same time and stores the final output in a huge memory space, `bottleneck reordering' simply changes the execution order of convolution in a channel-wise manner for consecutive convolution layers in the bottleneck, producing the exact same final output using much smaller (constant) memory.

\begin{figure}[!t]

  \centering
  \includegraphics[width=0.70\columnwidth]{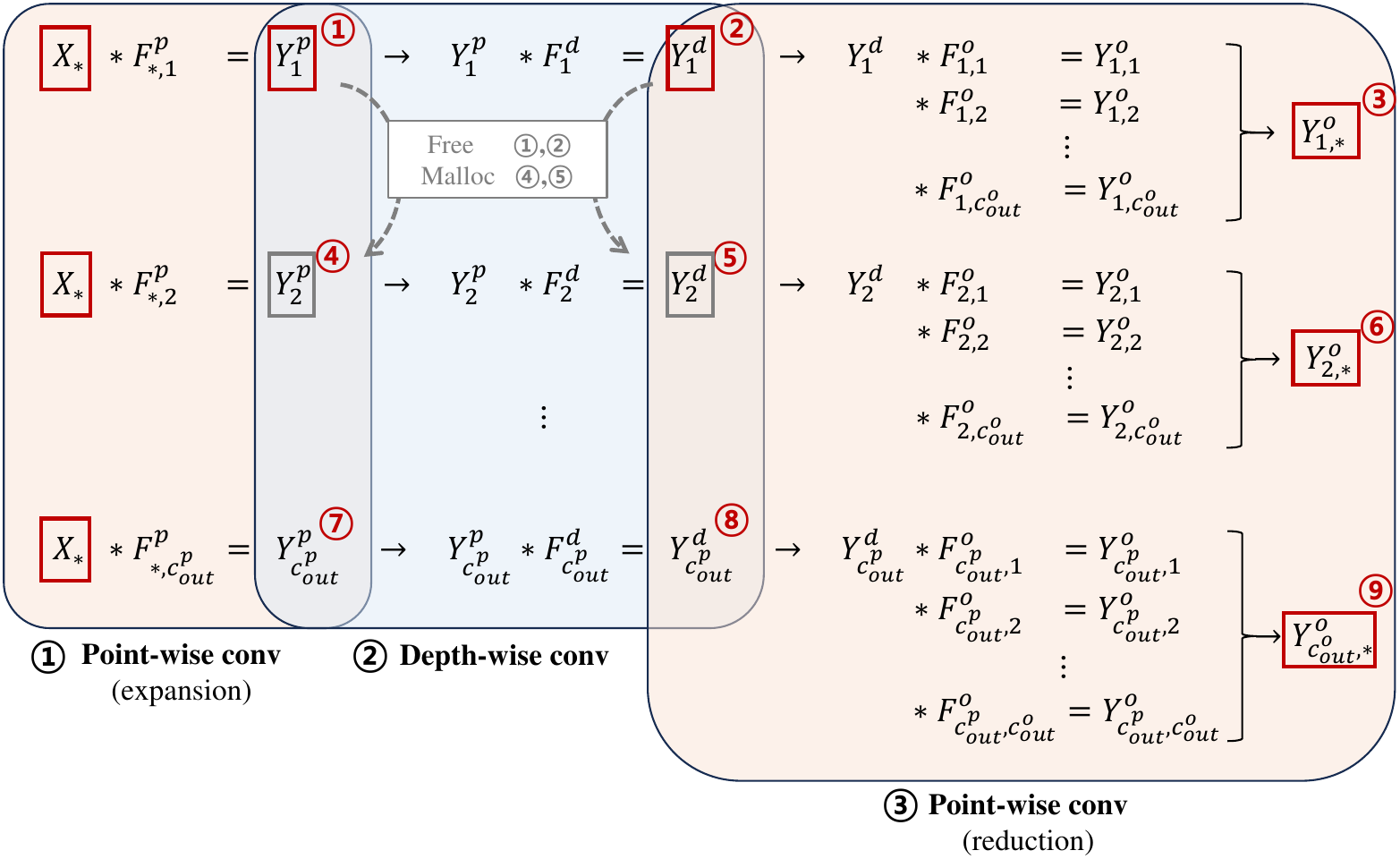}

  \caption{In a typical bottleneck operation, the black 1, 2, and 3 conv layers are processed sequentially. With bottleneck reordering, the layers are computed in the order of red (1, 2, 3), (4, 5, 6), and (7, 8, 9). After each cycle, two intermediate outputs are freed, and the next outputs (gray boxes and lines) are allocated. This approach reduces the peak memory requirement caused by intermediate outputs to \(1/c_{\text{out}}\).}
    \label{fig:reordering}

\end{figure}

By applying `bottleneck reordering' over three consecutive convolution layers in the bottleneck, \ie, the point-wise (expansion), depth-wise, and point-wise convolution (reduction), the total memory space for storing the related input and outputs of the bottleneck becomes at most the memory budget $m_{b}$ as:

\begin{equation}
    \underbrace{h_{i} \times w_{i} \times c_{i}}_{\substack{\text{Input of} \\ \text{bottleneck}}} + \underbrace{h_{o}^{p} \times w_{o}^{p} \times \mathcolorbox{light-gray}{1}}_{\substack{\text{Output of} \\ \text{point-wise conv}}} + \underbrace{h_{o}^{d} \times w_{o}^{d} \times \mathcolorbox{light-gray}{1}}_{\substack{\text{Output of} \\ \text{depth-wise conv}}} + \underbrace{h_{o} \times w_{o} \times c_{o}}_{\substack{\text{Output of} \\ \text{bottleneck}}} \leq m_{b}
    \label{eq:bottleneck}
\end{equation}
where $[h_{i} {\times} w_{i} {\times} c_{i}]$ and $[h_{o} {\times} w_{o} {\times} c_{o}]$ is the memory space for the initial input and the final output of the bottleneck, respectively, $[h_{o}^{p} {\times} w_{o}^{p} {\times} c_{out}^{p}]$ is the memory space for the output of the point-wise (expansion) conv, and $[h_{o}^{d} {\times} w_{o}^{d} {\times} c_{out}^{d}]$ is the memory space for the output of the depth-wise conv, with $c_{out}^{p} {=} c_{out}^{d} {=} 1$. \cref{fig:reordering} illustrates the reordering procedure, where $X$ is the input of the bottleneck, $F_{i}^{p}$, $F_{i}^{d}$, $F_{i}^{o}$ is the $i$-th kernel (filter) of the point-wise (expansion), depth-wise, and point-wise (reduction) convolution, $Y_{j}^{p}$, $Y_{j}^{d}$, $Y_{j}^{o}$ is the $j$-th channel output of the point-wise (expansion), depth-wise convolution, and bottleneck, respectively.

Note that the output channels of both the point-wise (expansion) and depth-wise convolution (\ie, $[h_{o}^{p} \times w_{o}^{p} \times 1]$ and $[h_{o}^{d} \times w_{o}^{d} \times 1]$) stays as one, \ie, $c_{out}^{p} = c_{out}^{d} = 1$, as highlighted as light-gray in \cref{eq:bottleneck}, meaning that the required output memory space does not increase over the number of output channels, \ie, $c_{out}^{p}$ and $c_{out}^{d}$, at all.  Considering that the point-wise (expansion) convolution of the bottleneck block is usually executed with a large size of output channels, \eg, $c_{out}^{p} = 100, 200, ...$, to improve the model performance, the proposed `bottleneck reordering' provides an effective way of constraining the large memory space for storing the output channels to a constant, regardless of the size of $c_{out}^{p}$ and $c_{out}^{d}$.

\section{Experiment}
\label{sec:experiment}
We implement a series of CNNs based on the proposed three design principles in C language and deploy them onto STM32 development boards~\cite{STM32H7}, featuring an ARM Cortex-M7 processor and 256 to 4,096 KB of SRAM. The networks are evaluated on two key tasks: image classification~\cite{deng2009imagenet,chowdhery2019visual} and object detection~\cite{Everingham2015,lin2015microsoft}.\\
\parlabel{Datasets.}
For image classification, we evaluate the proposed networks with ImageNet~\cite{deng2009imagenet} and VWW (Visual Wake Words)~\cite{chowdhery2019visual}. For object detection, we use PASCAL-VOC~\cite{Everingham2015} and MS-COCO~\cite{lin2015microsoft} for evaluation. \\
\parlabel{Network Architecture.}
By using the three proposed design principles (\cref{sec:method}), we construct CNNs with different memory budgets (\ie, $m_{b}$), ranging from 70 to 512 KB (float32) for image classification and from 259 to 1,719 KB (float32) for object detection, and then apply int8 quantization (Quant) to each of them. For image classification, the input image is segmented into four patches ($k{=}4$) along with the central patch, where each patch has $m_{h} {=} m_{w} {=} 18$ and $9$ of margins as the default setups, and experiments are conducted by changing them. For object detection, we attach the SSDLite~\cite{liu2016ssd} structure after bottleneck blocks.\\
\parlabel{Baselines.}
We compare the memory usage and model performance of the proposed networks with the two baselines, \ie, MobileNet~\cite{sandler2018mobilenetv2} and MCUNet~\cite{lin2020mcunet,lin2021mcunetv2}.
    \begin{table}[!t]
        \caption{The peak memory usage and classification accuracy (top-1 and top-5) of the proposed network on ImageNet~\cite{deng2009imagenet}, compared with MobileNet~\cite{sandler2018mobilenetv2} and MCUNet~\cite{lin2020mcunet,lin2021mcunetv2}.}
    \centering
        \begin{tabularx}{\textwidth}{l p{0.1cm} | p{0.1cm} rXcXcXcXcX}
        \toprule
        \textbf{Model Name} & & & \textbf{Peak-mem(KB)} & & \textbf{Quant} & & \textbf{Res} & & \textbf{Top-1(\%)} & & \textbf{Top-5(\%)} \\ \hline
        \multirow{2}{*}{\centering Proposed network} & & &  \makebox[0pt][r]{254/63}\hspace{0.8cm} & & fp32/int8 & & 224 & & 63.84/61.58 & & 84.80/83.26 \\
        & & &  \makebox[0pt][r]{533/133}\hspace{0.68cm} & & fp32/int8 & & 224 & & 64.68/61.50 & & 85.47/83.19 \\
        \hline
        \multirow{1}{*}{\centering MCUNetV1} & & &  \makebox[0pt][r]{968/242}\hspace{0.68cm} & & fp32/int8 & & 160 & & 60.90/60.30 & & 83.30/82.60 \\ \hline
        \multirow{2}{*}{\centering MCUNetV2} & & &  \makebox[0pt][r]{196}\hspace{1cm} & & int8 & & N/A & & 64.90 & & 86.20 \\
        & & &  \makebox[0pt][r]{456}\hspace{1cm} & & int8 & & N/A & & 71.80 & & 90.70 \\ \hline
        MobileNetV2 & & &  \makebox[0pt][r]{5,619}\hspace{0.9cm} & & fp32 & & 224 & & 72.83 & & 91.06 \\ \bottomrule
        \end{tabularx}
    \label{tab:experiment_overall_experiment}
    \end{table}
\begin{table}[!tb]
\caption{The peak memory usage and classification accuracy of the proposed network, MobileNet~\cite{sandler2018mobilenetv2}, and MCUNet~\cite{lin2020mcunet,lin2021mcunetv2} on VWW (Visual Wake Words)~\cite{chowdhery2019visual}.}

\centering
\begin{tabular}{l | r@{/}l  r@{/}l  c  r@{/}l }
    \toprule
    \textbf{Model Name} &   \multicolumn{2}{c}{\textbf{Peak-mem(KB)}} &   \multicolumn{2}{c}{\textbf{Quant}}  &   \textbf{Res}    &   \multicolumn{2}{c}{\textbf{Accuracy(\%)}} \\
    
    \midrule
    \multirow{2}{*}{\centering Proposed network}
        &   \makebox[1.2cm][r]{70}  &   \makebox[1.2cm][l]{18}  &   \makebox[1.2cm][r]{fp32}  &   \makebox[1.2cm][l]{int8}  &   \makebox[1.2cm][c]{66}    &   \makebox[1.2cm][r]{82.80}   &   \makebox[1.2cm][l]{82.63}   \\
        &   \makebox[1.2cm][r]{254} &   \makebox[1.2cm][l]{63}  &   \makebox[1.2cm][r]{fp32}  &   \makebox[1.2cm][l]{int8}  &   \makebox[1.2cm][c]{224}   &   \makebox[1.2cm][r]{90.67}   &   \makebox[1.2cm][l]{87.99}   \\
    \hline
    \multirow{3}{*}{MCUNetV1}
        &   \makebox[1.2cm][r]{584} &   \makebox[1.2cm][l]{146} &   \makebox[1.2cm][r]{fp32}  &   \makebox[1.2cm][l]{int8}  &   \makebox[1.2cm][c]{64}    &   \makebox[1.2cm][r]{87.40}   &   \makebox[1.2cm][l]{87.30}   \\
        &   \makebox[1.2cm][r]{648} &   \makebox[1.2cm][l]{162} &   \makebox[1.2cm][r]{fp32}  &   \makebox[1.2cm][l]{int8}  &   \makebox[1.2cm][c]{160}   &   \makebox[1.2cm][r]{88.90}   &   \makebox[1.2cm][l]{88.90}   \\
        &   \makebox[1.2cm][r]{1,244}&  \makebox[1.2cm][l]{311} &   \makebox[1.2cm][r]{fp32}  &   \makebox[1.2cm][l]{int8}  &   \makebox[1.2cm][c]{144}   &   \makebox[1.2cm][r]{91.70}   &   \makebox[1.2cm][l]{91.80}   \\
    \hline
    \multirow{3}{*}{MCUNetV2}
        &   \multicolumn{2}{c}{\makebox[1.2cm]{30}}  &   \multicolumn{2}{c}{\makebox[1.2cm]{int8}} & \makebox[1.2cm]{N/A} & \multicolumn{2}{c}{\makebox[1.2cm]{90.00}} \\
        &   \multicolumn{2}{c}{\makebox[1.2cm]{62}}  &   \multicolumn{2}{c}{\makebox[1.2cm]{int8}} & \makebox[1.2cm]{N/A} & \multicolumn{2}{c}{\makebox[1.2cm]{93.00}} \\
        &   \multicolumn{2}{c}{\makebox[1.2cm]{118}} &   \multicolumn{2}{c}{\makebox[1.2cm]{int8}} & \makebox[1.2cm]{N/A} & \multicolumn{2}{c}{\makebox[1.2cm]{94.00}} \\
    \hline
\end{tabular}
\label{tab:experiment_vww}
\end{table}
\subsection{Image Classification}
\parlabel{Memory Usage and Model Performance.}
\cref{tab:experiment_overall_experiment} and \ref{tab:experiment_vww} show the proposed networks with different memory budgets (usages) and classification accuracy on ImageNet~\cite{deng2009imagenet} and VWW~\cite{chowdhery2019visual}, compared with the two baselines, \ie, MobileNet~\cite{sandler2018mobilenetv2} and MCUNet~\cite{lin2020mcunet,lin2021mcunetv2}. In \cref{tab:experiment_overall_experiment}, the proposed network takes the lowest memory among all, \ie, 63 KB with quantization (int8), 3.1x smaller than the smallest one of the baselines (196 KB), achieving 61.58\% and 83.26\% top-1 and top-5 accuracy. Even without quantization, it can be seen that the proposed network still takes the relatively lower memory, \ie, 254 KB (float32), achieving 63.84\% top-1 accuracy, comparable to MCUNet requiring 196 KB with quantization (int8) to provide 64.90\% top-1 accuracy. In \cref{tab:experiment_vww}, it also takes the smallest memory, \ie, 18 KB, while achieving comparable classification accuracy, \ie, 82.63\%. It shows that the proposed three design principles allow for flexibly building memory-efficient networks that fits extremely tight memory budgets in a memory-aware manner while providing comparable model performance.

\parlabel{Channel Output.}
\cref{tab:experiment_proposed_model_t} shows the peak memory usage and classification accuracy of ImageNet~\cite{deng2009imagenet} with different numbers of expansion ratio, \ie, $t$, of the point-wise (expansion) convolution in the bottleneck. As shown in the table, the peak memory usage does not increase over $t$, enabled by `bottleneck reordering', whereas the classification accuracy increases, \eg, 58.82\% ($t {=} 2$) and 63.84\% ($t {=} 8$) top-1 accuracy, entailing a slight increase in the point-wise kernel size. It shows that a flexible trade-off exists between the model performance and kernel size, whereas the peak memory usage remains constant.
\begin{table}[!tb]
\caption{The peak memory usage and classification accuracy on ImageNet~\cite{deng2009imagenet} with various expansion ratios (\ie, \(t\)) in the bottleneck. The value of \(c_{out}\), representing the number of output channels after channel expansion, is derived by multiplying \(t\) with the number of output channels from the preceding depthwise convolution layer.
}
\centering
\begin{tabular}{l | r@{/}l  r@{/}l  r@{/}l r@{/}l}
    \toprule
    \textbf{\(\mathbf{t}\)} &   \multicolumn{2}{c}{\textbf{Peak-mem(KB)}} &   \multicolumn{2}{c}{\textbf{Quant}}  &   \multicolumn{2}{c}{\textbf{Top-1(\%)}}    &   \multicolumn{2}{c}{\textbf{Top-5(\%)}} \\
    \midrule
    2   &   \makebox[1.2cm][r]{254}  &   \makebox[1.2cm][l]{63}&   \makebox[1.2cm][r]{fp32} &\makebox[1.2cm][l]{int8}  &   \makebox[1.2cm][r]{58.82}&   \makebox[1.2cm][l]{55.75}  &   \makebox[1.2cm][r]{81.04}&   \makebox[1.2cm][l]{78.97}     \\
    4   &   \makebox[1.2cm][r]{254}  &   \makebox[1.2cm][l]{63}&   \makebox[1.2cm][r]{fp32} &\makebox[1.2cm][l]{int8}  &   \makebox[1.2cm][r]{62.28}&   \makebox[1.2cm][l]{58.98}  &   \makebox[1.2cm][r]{83.76}&   \makebox[1.2cm][l]{81.63}     \\
    6   &   \makebox[1.2cm][r]{254}  &   \makebox[1.2cm][l]{63}&   \makebox[1.2cm][r]{fp32} &\makebox[1.2cm][l]{int8}  &   \makebox[1.2cm][r]{63.48}&   \makebox[1.2cm][l]{60.70}  &   \makebox[1.2cm][r]{84.98}&   \makebox[1.2cm][l]{82.92}     \\
    8   &   \makebox[1.2cm][r]{254}  &   \makebox[1.2cm][l]{63}&   \makebox[1.2cm][r]{fp32} &\makebox[1.2cm][l]{int8}  &   \makebox[1.2cm][r]{63.84}&   \makebox[1.2cm][l]{61.58}  &   \makebox[1.2cm][r]{84.80}&   \makebox[1.2cm][l]{83.26}     \\
    \bottomrule
\end{tabular}
\label{tab:experiment_proposed_model_t}
\end{table}
\parlabel{Patch Composition.}
\cref{tab:experiment_no_center_patch} shows the peak memory usage and classification accuracy of ImageNet~\cite{deng2009imagenet} and VWW~\cite{chowdhery2019visual} when removing the central patch tunnel from the network. As shown in the table, the classification accuracy drops for both ImageNet (from 63.84\% to 62.32\%) and VWW (from 82.80\% to 78.97\%), substantiating the crucial role of the central patch in classification.
\noindent \cref{{tab:experiment_different_patches}} shows the peak memory usage and classification accuracy of ImageNet~\cite{deng2009imagenet} and VWW~\cite{chowdhery2019visual} over different numbers of patches (\ie, from $k=5$ to $k=2$), which learns the non-segmented entire images resized to $[130 {\times} 130 {\times} 3]$ and $[42 {\times} 42{\times} 3]$, respectively, where the two top rows show the cases when learning the proposed five segmented image patches (\ie, `input segmentation'). It shows that `input segmentation' helps boost the model performance, while learning the same entire images as multiple patches does not provide similar performance regardless of the number of patches (\eg, 63.84\% vs. 60.70\% in ImageNet~\cite{deng2009imagenet}), demonstrating the proposed `input segmentation' is essential to achieve competitive performance.
\begin{table}[!tb]
\caption{The experiments on ImageNet~\cite{deng2009imagenet} and VWW (Visual Wake Words)~\cite{chowdhery2019visual} without the central patch (\ie, having only four patches in the network).}
\centering
\begin{tabularx}{\textwidth}{cXcXrXcXcXcX}
\toprule
\textbf{Dataset} && \textbf{\# of Patches} && \textbf{Peak-mem(KB)} && \textbf{Quant} && \textbf{Top-1(\%)} && \textbf{Top-5(\%)} \\ \midrule
ImageNet && 4    && \makebox[0pt][r]{254/63}\hspace{1cm} && fp32/int8  && 62.32/84.38     && 84.38  \\ \hline
VWW      && 4    && \makebox[0pt][r]{70/18}\hspace{1cm}  && fp32/int8  && 78.97/78.46 && N/A   \\ \bottomrule
\end{tabularx}
\label{tab:experiment_no_center_patch}
\end{table}
\begin{table}[!tb]
    \caption{The experiments on ImageNet~\cite{deng2009imagenet} and VWW (Visual Wake Words)~\cite{chowdhery2019visual} with different numbers of patches (\ie, $k$). The top two rows show the result of the segmented patches, while the remaining rows show the result of resizing the non-segmented images.}
    \begin{tabular}{@{}p{0.48\textwidth}@{} p{0.04\textwidth} @{}p{0.48\textwidth}@{}}
        \begin{minipage}[t]{0.48\textwidth}
            \begin{tabularx}{\textwidth}{>{\centering\arraybackslash}X >{\centering\arraybackslash}X >{\centering\arraybackslash}X >{\centering\arraybackslash}X >{\centering\arraybackslash}X}
                \toprule
                \multicolumn{5}{c}{\textbf{ImageNet}} \\ \midrule
                \textbf{Patch} & \textbf{Mem} & \textbf{Quant} & \textbf{Top-1} & \textbf{Top-5} \\ \midrule
                5 & \hspace{1cm}\makebox[0pt][r]{254KB}\hspace{0.05cm} & fp32  & 63.84\%  & 84.80\% \\ 
                5 & \hspace{1cm}\makebox[0pt][r]{63KB}\hspace{0.05cm} & int8  & 61.58\%  & 83.26\%  \\ \hline
                5 & \hspace{1cm}\makebox[0pt][r]{254KB}\hspace{0.05cm} & fp32  & 60.70\% & 82.59\%\ \\ 
                5 & \hspace{1cm}\makebox[0pt][r]{63KB}\hspace{0.05cm} & int8  & 57.61\%  & 80.80\%       \\ 
                4 & \hspace{1cm}\makebox[0pt][r]{254KB}\hspace{0.05cm} & fp32  & 60.83\% & 82.93\%\\ 
                4 & \hspace{1cm}\makebox[0pt][r]{63KB}\hspace{0.05cm} & int8  & 57.65\%     & 80.16\%       \\ 
                3 & \hspace{1cm}\makebox[0pt][r]{254KB}\hspace{0.05cm} & fp32  & 59.14\%  & 81.90\% \\ 
                3 & \hspace{1cm}\makebox[0pt][r]{63KB}\hspace{0.05cm} & int8  & 58.54\%   & 81.85\% \\ 
                2 & \hspace{1cm}\makebox[0pt][r]{254KB}\hspace{0.05cm} & fp32  & 57.13\%  & 80.40\% \\ 
                2 & \hspace{1cm}\makebox[0pt][r]{63KB}\hspace{0.05cm} & int8  & 55.67\%   & 78.97\% \\ \bottomrule
            \end{tabularx}
        \end{minipage} & &
        \begin{minipage}[t]{0.48\textwidth}
            \begin{tabularx}{\textwidth}{>{\centering\arraybackslash}X >{\centering\arraybackslash}X >{\centering\arraybackslash}X >{\centering\arraybackslash}X}
                \toprule
                \multicolumn{4}{c}{\textbf{VWW}} \\ \midrule
                \textbf{Patch} & \textbf{Mem} & \textbf{Quant} & \textbf{Top-1}\\ \midrule
                5 & \hspace{0.8cm}\makebox[0pt][r]{70KB}\hspace{0.05cm} & fp32  & 82.80\%  \\ 
                5 & \hspace{0.8cm}\makebox[0pt][r]{18KB}\hspace{0.05cm} & int8  & 82.63\% \\ \hline
                5 & \hspace{0.8cm}\makebox[0pt][r]{70KB}\hspace{0.05cm} & int8  & 81.56\% \\ 
                5 & \hspace{0.8cm}\makebox[0pt][r]{18KB}\hspace{0.05cm} & fp32  & 81.39\%  \\
                4 & \hspace{0.8cm}\makebox[0pt][r]{70KB}\hspace{0.05cm} & fp32  & 81.46\% \\ 
                4 & \hspace{0.8cm}\makebox[0pt][r]{18KB}\hspace{0.05cm} & int8  & 81.02\% \\ 
                3 & \hspace{0.8cm}\makebox[0pt][r]{70KB}\hspace{0.05cm} & fp32  & 81.23\%\\ 
                3 & \hspace{0.8cm}\makebox[0pt][r]{18KB}\hspace{0.05cm} & int8  & 80.97\%\\ 
                2 & \hspace{0.8cm}\makebox[0pt][r]{70KB}\hspace{0.05cm} & fp32  & 81.10\%\\ 
                2 & \hspace{0.8cm}\makebox[0pt][r]{18KB}\hspace{0.05cm} & int8  & 80.12\%\\ \bottomrule
            \end{tabularx}
        \end{minipage}
    \end{tabular}
    \label{tab:experiment_different_patches}
\end{table}
\begin{table}[!tb]
    \caption{The experiments on ImageNet~\cite{deng2009imagenet} and VWW (Visual Wake Words)~\cite{chowdhery2019visual} by varying the patch margin (\ie, $m_h$ and $m_w$).}
    \begin{minipage}[t]{0.48\textwidth}
        \begin{tabularx}{\textwidth}{>{\centering\arraybackslash}X >{\centering\arraybackslash}X >{\centering\arraybackslash}X >{\centering\arraybackslash}X >{\centering\arraybackslash}X}
            \toprule
            \multicolumn{5}{c}{\textbf{ImageNet}} \\ \midrule
            \textbf{Margin} & \textbf{Mem} & \textbf{Quant} & \textbf{Top-1} & \textbf{Top-5} \\ \midrule
            18 & \hspace{1cm}\makebox[0pt][r]{254KB} & fp32  & 63.84\%  & 84.80\% \\ 
            18 & \hspace{1cm}\makebox[0pt][r]{63KB} & int8  & 61.58\%  & 83.26\% \\ \hline
            10 & \hspace{1cm}\makebox[0pt][r]{208KB} & fp32  & 62.23\%  & 83.85\% \\ 
            10 & \hspace{1cm}\makebox[0pt][r]{52KB} & int8  & 60.00\%  & 82.42\% \\ 
            5  & \hspace{1cm}\makebox[0pt][r]{195KB} & fp32  & 62.12\%  & 83.91\% \\ 
            5  & \hspace{1cm}\makebox[0pt][r]{49KB} & int8  & 59.51\%  & 82.25\% \\ 
            0  & \hspace{1cm}\makebox[0pt][r]{175KB} & fp32  & 61.31\%  & 83.39\% \\ 
            0  & \hspace{1cm}\makebox[0pt][r]{44KB} & int8  & 58.33\%  & 81.33\% \\ \bottomrule
        \end{tabularx}
    \end{minipage} 
    \hfill
    \begin{minipage}[t]{0.48\textwidth}
        \begin{tabularx}{\textwidth}{>{\centering\arraybackslash}X >{\centering\arraybackslash}X >{\centering\arraybackslash}X >{\centering\arraybackslash}X}
            \toprule
            \multicolumn{4}{c}{\textbf{VWW}} \\ \midrule
            \textbf{Margin} & \textbf{Mem} & \textbf{Quant} & \textbf{Top-1} \\ \midrule
            9  & \hspace{0.8cm}\makebox[0pt][r]{70KB}\hspace{0.1cm}  & fp32 & 82.80\% \\ 
            9  & \hspace{0.8cm}\makebox[0pt][r]{18KB}\hspace{0.1cm}  & int8 & 82.63\% \\ \hline
            5  & \hspace{0.8cm}\makebox[0pt][r]{70KB}\hspace{0.1cm}  & fp32 & 80.48\% \\ 
            5  & \hspace{0.8cm}\makebox[0pt][r]{18KB}\hspace{0.1cm}  & int8 & 80.79\%\\ 
            2  & \hspace{0.8cm}\makebox[0pt][r]{70KB}\hspace{0.1cm}  & fp32 & 80.05\% \\ 
            2  & \hspace{0.8cm}\makebox[0pt][r]{18KB}\hspace{0.1cm}  & int8 & 79.65\%\\ 
            0  & \hspace{0.8cm}\makebox[0pt][r]{70KB}\hspace{0.1cm}  & fp32 & 80.00\% \\ 
            0  & \hspace{0.8cm}\makebox[0pt][r]{18KB}\hspace{0.1cm}  & int8 & 79.60\%\\ \bottomrule
        \end{tabularx}
    \end{minipage}
    \label{tab:experiment_different_overlap_size}
\end{table}
\parlabel{Margin Size.}
\cref{tab:experiment_different_overlap_size} shows the peak memory usage and classification accuracy of ImageNet~\cite{deng2009imagenet} and VWW~\cite{chowdhery2019visual} over different sizes of margins, \ie, $m_{h}$ and $m_{w}$, where larger margins tend to achieve better model performances in general. It demonstrates that adding some margin to the patch, leading them to slightly overlap with each other, is crucial to the model performance. However, since a larger margin (\ie, $m_{h}$ and $m_{w}$) increases the memory requirement as in \cref{eq:margin}, the proper margin values should be chosen based on the memory budget.
\subsection{Object Detection}
\begin{figure} [!tb]
  \centering
  \includegraphics[width=0.5\textwidth]{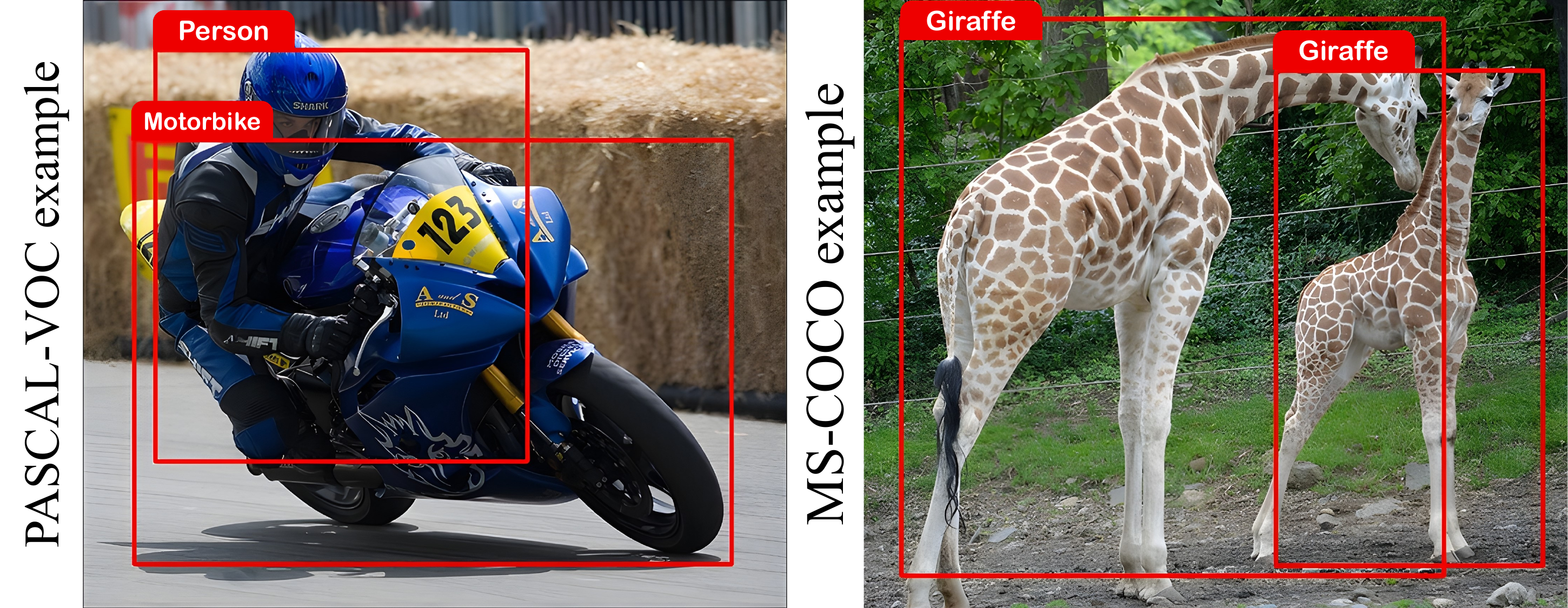}
  \caption{Examples of object detection performed by the proposed network: (Left) PASCAL-VOC~\cite{Everingham2015} and (Right) MS-COCO~\cite{lin2015microsoft}.}
    \label{fig:detection_examples}
\end{figure}

\parlabel{Memory Usage and Model Performance.}
\cref{tab:experiment_detection_overall_voc} and \ref{tab:experiment_detection_overall_coco} show the proposed networks with different memory budgets and mAP (mean Average Precision) on PASCAL-VOC~\cite{Everingham2015} and MS-COCO~\cite{lin2015microsoft}, compared with MobileNet~\cite{sandler2018mobilenetv2} and MCUNet~\cite{lin2020mcunet,lin2021mcunetv2}. For MS-COCO, all networks are trained with trainval35k 
and evaluated on minival5k in the MS-COCO 2014 dataset. Similar to image classification, the proposed network takes the lowest memory among all in \cref{tab:experiment_detection_overall_voc}, \ie, 65 KB with quantization (int8), 155x and 3.8x smaller than MobileNet (10,080 KB) and MCUNet (247 KB), respectively, achieving 50.7\% mAP, showing that it effectively trades off the peak memory usage and mAP. However, the network taking the second lowest memory, \ie, 176 KB, achieves 62.1\% mAP, comparable to MCUNet taking 247 KB to provide 64.6\% mAP. In \cref{tab:experiment_detection_overall_coco}, the proposed network also takes much smaller memory compared to MobileNet, \eg, 1,719 KB (5.8x smaller) with a reasonable level of mAP (30.0\%), verifying that the proposed design principles also effectively apply to object detection. \cref{fig:detection_examples} shows examples of object detection on PASCAL-VOC~\cite{Everingham2015} and MS-COCO.
\begin{table}[!tb]
\caption{The peak memory usage and mAP (mean Average Precision) of the proposed network, MobileNet~\cite{sandler2018mobilenetv2}, and MCUNet~\cite{lin2020mcunet,lin2021mcunetv2} on PASCAL-VOC~\cite{Everingham2015}.}
\centering
\begin{tabular}{l | r@{/}l  r@{/}l  c  r@{/}l }
    \toprule
    \textbf{Model Name} &   \multicolumn{2}{c}{\textbf{Peak-mem(KB)}} &   \multicolumn{2}{c}{\textbf{Quant}}  &   \textbf{Res}    &   \multicolumn{2}{c}{\textbf{mAP (\%)}} \\
    \midrule
    \multirow{3}{*}{\centering Proposed network}
        &   \makebox[1.2cm][r]{259}  &   \makebox[1.2cm][l]{65}  &   \makebox[1.2cm][r]{fp32}  &   \makebox[1.2cm][l]{int8}  &   \makebox[1.2cm][c]{130}    &   \makebox[1.2cm][r]{51.8}   &   \makebox[1.2cm][l]{50.7}   \\
        &   \makebox[1.2cm][r]{702} &   \makebox[1.2cm][l]{176}  &   \makebox[1.2cm][r]{fp32}  &   \makebox[1.2cm][l]{int8}  &   \makebox[1.2cm][c]{224}   &   \makebox[1.2cm][r]{63.4}   &   \makebox[1.2cm][l]{62.1}   \\
        &   \makebox[1.2cm][r]{1,260} &   \makebox[1.2cm][l]{315}  &   \makebox[1.2cm][r]{fp32}  &   \makebox[1.2cm][l]{int8}  &   \makebox[1.2cm][c]{300}   &   \makebox[1.2cm][r]{67.0}   &   \makebox[1.2cm][l]{66.3}   \\
    \hline
    \multirow{1}{*}{MCUNetV1}
        &   \multicolumn{2}{c}{\makebox[1.2cm]{466}}  &   \multicolumn{2}{c}{\makebox[1.2cm]{int8}} & \makebox[1.2cm]{224} & \multicolumn{2}{c}{\makebox[1.2cm]{51.4}} \\
    \hline
    \multirow{2}{*}{MCUNetV2}
        &   \multicolumn{2}{c}{\makebox[1.2cm]{438}}  &   \multicolumn{2}{c}{\makebox[1.2cm]{int8}} & \makebox[1.2cm]{224} & \multicolumn{2}{c}{\makebox[1.2cm]{68.3}} \\
        &   \multicolumn{2}{c}{\makebox[1.2cm]{247}}  &   \multicolumn{2}{c}{\makebox[1.2cm]{int8}} & \makebox[1.2cm]{224} & \multicolumn{2}{c}{\makebox[1.2cm]{64.6}} \\
    \hline
    \multirow{1}{*}{MobileNetV2}
        &   \multicolumn{2}{c}{\makebox[1.2cm]{10,080}}  &   \multicolumn{2}{c}{\makebox[1.2cm]{fp32}} & \makebox[1.2cm]{300} & \multicolumn{2}{c}{\makebox[1.2cm]{68.6}} \\
    \hline
\end{tabular}
\label{tab:experiment_detection_overall_voc}
\end{table}
\begin{table}[!tb]
\caption{The peak memory usage and mAP (mean Average Precision) of the proposed network, MobileNet~\cite{sandler2018mobilenetv2}, and MCUNet~\cite{lin2020mcunet,lin2021mcunetv2} on MS-COCO~\cite{lin2015microsoft}.}
\centering
\begin{tabular}{l | r@{/}l  r@{/}l  c  r@{/}l }
    \toprule
    \textbf{Model Name} &   \multicolumn{2}{c}{\textbf{Peak-mem(KB)}} &   \multicolumn{2}{c}{\textbf{Quant}}  &   \textbf{Res}    &   \multicolumn{2}{c}{\textbf{\(\mathbf{mAP^{val}}\)(\%)}} \\
    \midrule
    \multirow{3}{*}{\centering Proposed network}
        &   \makebox[1.2cm][r]{446}  &   \makebox[1.2cm][l]{112}  &   \makebox[1.2cm][r]{fp32}  &   \makebox[1.2cm][l]{int8}  &   \makebox[1.2cm][c]{130}    &   \makebox[1.2cm][r]{22.0}   &   \makebox[1.2cm][l]{21.2}   \\
        &   \makebox[1.2cm][r]{1,210} &   \makebox[1.2cm][l]{303}  &   \makebox[1.2cm][r]{fp32}  &   \makebox[1.2cm][l]{int8}  &   \makebox[1.2cm][c]{224}   &   \makebox[1.2cm][r]{28.9}   &   \makebox[1.2cm][l]{27.2}   \\
        &   \makebox[1.2cm][r]{1,719} &   \makebox[1.2cm][l]{430}  &   \makebox[1.2cm][r]{fp32}  &   \makebox[1.2cm][l]{int8}  &   \makebox[1.2cm][c]{224}   &   \makebox[1.2cm][r]{30.0}   &   \makebox[1.2cm][l]{28.0}   \\
    \hline
    \multirow{1}{*}{MobileNetV2}
        &   \multicolumn{2}{c}{\makebox[1.2cm]{10,080}}  &   \multicolumn{2}{c}{\makebox[1.2cm]{fp32}} & \makebox[1.2cm]{300} & \multicolumn{2}{c}{\makebox[1.2cm]{31.8}} \\
    \hline
\end{tabular}
\label{tab:experiment_detection_overall_coco}
\end{table}

\parlabel{Channel Output.}
\cref{tab:experiment_detection_cout} shows the peak memory usage and mAP (mean Average Precision) on PASCAL-VOC and MS-COCO with different numbers of expansion ratio, \ie, $t$, of the point-wise (expansion) convolution in the bottleneck. As shown in the table, the peak memory usage does not increase over $t$, enabled by `bottleneck reordering', whereas mAP increase, \eg, from 52.3\% to 63.4\% in PASCAL-VOC, along with a slight increase in the convolution kernel size. Similar to the image classification tasks, the proposed network provides a flexible trade-off between mAP and the size of the kernel for object detection, while keeping the peak memory requirement constant.
\begin{table} [!tb]
    \caption{The object detection on PASCAL-VOC~\cite{Everingham2015} and MS-COCO~\cite{lin2015microsoft} with different expansion ratios (\ie, \(t\)) in the point-wise (expansion) conv within the bottleneck.}
    \centering
    \begin{tabular}{l | r@{/}l  r@{/}l  c  r@{/}l || r@{/}l  r@{/}l  c  r@{/}l }
        \toprule
        \multicolumn{8}{c||}{\textbf{Pascal-VOC}}   & \multicolumn{7}{c}{\textbf{MS-COCO}} \\
        \hline
        \textbf{\(\mathbf{t}\)} &   \multicolumn{2}{c}{\textbf{Mem(KB)}} &   \multicolumn{2}{c}{\textbf{Quant}}  &   \textbf{Res}    &   \multicolumn{2}{c||}{\textbf{mAP(\%)}} &   \multicolumn{2}{c}{\textbf{Mem(KB)}} &   \multicolumn{2}{c}{\textbf{Quant}}  &   \textbf{Res}    &   \multicolumn{2}{c}{\textbf{mAP(\%)}} \\
        \midrule
        2 & \makebox[0.8cm][r]{446} & \makebox[0.8cm][l]{112} & fp32  & int8  & \makebox[1cm][c]{130} & 22.0   & 21.2 & \makebox[0.8cm][r]{446}  & \makebox[0.8cm][l]{112}  & fp32  & int8  & \makebox[1cm][c]{130} & 22.0   & 21.2  \\
        2 & \makebox[0.8cm][r]{210} & \makebox[0.8cm][l]{303} & fp32  & int8  & \makebox[1cm][c]{224} & 28.9   & 27.2 & \makebox[0.8cm][r]{446}  & \makebox[0.8cm][l]{112}  & fp32  & int8  & \makebox[1cm][c]{130} & 22.0   & 21.2  \\
        2 & \makebox[0.8cm][r]{719} & \makebox[0.8cm][l]{430} & fp32  & int8  & \makebox[1cm][c]{224} & 30.0   & 28.0 & \makebox[0.8cm][r]{446}  & \makebox[0.8cm][l]{112}  & fp32  & int8  & \makebox[1cm][c]{130} & 22.0   & 21.2 \\
        \hline
    \end{tabular}
    \label{tab:experiment_detection_cout}
\end{table}

\section{Limitations and Discussions}

\parlabel{Network Size.} Although the proposed network takes much smaller memory when compared to state-of-the-art memory-efficient MCUNet~\cite{lin2020mcunet,lin2021mcunetv2} and MobileNet~\cite{sandler2018mobilenetv2}, \eg, 89x and 3.1x smaller for ImageNet~\cite{deng2009imagenet}, respectively, the network size is not decreased significantly, unlike huge memory reduction. That is because the number of weight parameters of the proposed network for ImageNet, \ie, 3.34 million, is similar to that of MobileNet (\ie 3.2 million) and larger than that of MCUNet (\ie 0.73 million). It can be regarded as a trade-off between memory usage and network size, which can be flexibly adjusted based on the system and/or application requirements. Since the MCUNet architecture was designed using NAS, unlike our network, we expect that NAS could identify a more efficient architecture in terms of weight parameters while maintaining our three proposed design principles.
Alternatively, the weight parameters of $k$ patch tunnels could be shared among them to reduce the total number of weight parameters of the network by $k$ times. Also, it can be considered to combine some patch tunnels at the deeper layer of the network where the memory usage is smaller, as seen in \cref{fig:memory}. We will investigate these approaches in future work. \\
\parlabel{Larger Images.} The proposed `input segmentation' and `patch tunneling', in principle, can be applied to images larger than the ImageNet data (\ie $224 \times 224$) by adjusting the number of patches and the overlapped areas accordingly based on the memory budget. However, as a large image is segmented into an increased number of smaller patches, the amount of information included in a single patch tends to decrease, making it more difficult to look at and learn the image from a global perspective. As each patch tunnel separately learns features available only in an individual local patch, the feature aggregation procedure for the global image learning performed at the last layer of the network is likely to become more difficult, possibly causing model performance degradation. We expect that it can be mitigated by segmenting an image into patches more flexibly, such as making different sizes of patches at various positions of the image which can cover different areas of the image with varying sizes. We also leave it for future work.\\
\parlabel{Other Tasks and Datasets}: In this work, we construct memory-efficient CNNs for two vision tasks, \ie, image classification and object detection, demonstrating its effectiveness. For a more thorough evaluation, however, additional experiments on different tasks and various datasets, such as image segmentation~\cite{cordts2016cityscapesdatasetsemanticurban,zhou2018semanticunderstandingscenesade20k}, super resolution~\cite{magnitskaya2017electronphononpropertiesnoncentrosymmetric,dong2015imagesuperresolutionusingdeep}, and image generation{~\cite{liu2015deeplearningfaceattributes, yu2016lsunconstructionlargescaleimage}}, should be conducted. 

\section{Conclusion}

We introduce a bottleneck-based, extremely memory-efficient CNN that fits the tight memory constraints (\eg, under 256 KB) of embedded and IoT devices. The network is built on three design principles: input segmentation', patch tunneling', and `bottleneck reordering', allowing flexible adjustment of memory usage based on device capacity. We implement the network on an actual device (STM32H7~\cite{STM32H7}), demonstrating that these design principles enable memory-aware CNNs for various vision tasks, such as memory-efficient image classification on ImageNet~\cite{deng2009imagenet} using 63 KB (61.58\% top-1 accuracy) and object detection on PASCAL-VOC~\cite{Everingham2015} using 65 KB (50.7\% mAP).

\section*{Acknowledgement}
This research was supported by the Challengeable Future Defense Technology Research and Development Program through the Agency for Defense Development(ADD) funded by the Defense Acquisition Program Administration(DAPA) in 2022(No.915062201), and Institute of Information \& communications Technology Planning \& Evaluation(IITP) grant funded by the Korea government(MSIT) (No.RS-2020-II201336, Artificial Intelligence Graduate School Program(UNIST)). 

\bibliographystyle{splncs04}
\bibliography{main}

\begin{thebibliography}{10}
\providecommand{\url}[1]{\texttt{#1}}
\providecommand{\urlprefix}{URL }
\providecommand{\doi}[1]{https://doi.org/#1}

\bibitem{bigas2006review}
Bigas, M., Cabruja, E., Forest, J., Salvi, J.: Review of cmos image sensors. Microelectronics journal  \textbf{37}(5),  433--451 (2006)

\bibitem{choi2015always}
Choi, J., Shin, J., Kang, D., Park, D.S.: Always-on cmos image sensor for mobile and wearable devices. IEEE Journal of Solid-State Circuits  \textbf{51}(1),  130--140 (2015)

\bibitem{chowdhery2019visual}
Chowdhery, A., Warden, P., Shlens, J., Howard, A., Rhodes, R.: Visual wake words dataset (2019)

\bibitem{cordts2016cityscapesdatasetsemanticurban}
Cordts, M., Omran, M., Ramos, S., Rehfeld, T., Enzweiler, M., Benenson, R., Franke, U., Roth, S., Schiele, B.: The cityscapes dataset for semantic urban scene understanding (2016), \url{https://arxiv.org/abs/1604.01685}

\bibitem{deng2009imagenet}
Deng, J., Dong, W., Socher, R., Li, L.J., Li, K., Fei-Fei, L.: Imagenet: A large-scale hierarchical image database. In: 2009 IEEE conference on computer vision and pattern recognition. pp. 248--255. Ieee (2009)

\bibitem{dhar2021survey}
Dhar, S., Guo, J., Liu, J., Tripathi, S., Kurup, U., Shah, M.: A survey of on-device machine learning: An algorithms and learning theory perspective. ACM Transactions on Internet of Things  \textbf{2}(3),  1--49 (2021)

\bibitem{dong2015imagesuperresolutionusingdeep}
Dong, C., Loy, C.C., He, K., Tang, X.: Image super-resolution using deep convolutional networks (2015), \url{https://arxiv.org/abs/1501.00092}

\bibitem{elsken2019neural}
Elsken, T., Metzen, J.H., Hutter, F.: Neural architecture search: A survey. The Journal of Machine Learning Research  \textbf{20}(1),  1997--2017 (2019)

\bibitem{Everingham2015}
Everingham, M., Eslami, S.M.A., Van~Gool, L., Williams, C.K.I., Winn, J., Zisserman, A.: The pascal visual object classes challenge: A retrospective. International Journal of Computer Vision  \textbf{111}(1),  98--136 (Jan 2015). \doi{10.1007/s11263-014-0733-5}, \url{https://doi.org/10.1007/s11263-014-0733-5}

\bibitem{gholami2022survey}
Gholami, A., Kim, S., Dong, Z., Yao, Z., Mahoney, M.W., Keutzer, K.: A survey of quantization methods for efficient neural network inference. In: Low-Power Computer Vision, pp. 291--326. Chapman and Hall/CRC (2022)

\bibitem{han2015deep}
Han, S., Mao, H., Dally, W.J.: Deep compression: Compressing deep neural networks with pruning, trained quantization and huffman coding. arXiv preprint arXiv:1510.00149  (2015)

\bibitem{he2021automl}
He, X., Zhao, K., Chu, X.: Automl: A survey of the state-of-the-art. Knowledge-Based Systems  \textbf{212},  106622 (2021)

\bibitem{huang2019efficient}
Huang, K., Ni, B., Yang, X.: Efficient quantization for neural networks with binary weights and low bitwidth activations. In: Proceedings of the AAAI Conference on Artificial Intelligence. vol.~33, pp. 3854--3861 (2019)

\bibitem{liang2021pruning}
Liang, T., Glossner, J., Wang, L., Shi, S., Zhang, X.: Pruning and quantization for deep neural network acceleration: A survey. Neurocomputing  \textbf{461},  370--403 (2021)

\bibitem{liberis2021munas}
Liberis, E., Dudziak, {\L}., Lane, N.D.: $\mu$nas: Constrained neural architecture search for microcontrollers. In: Proceedings of the 1st Workshop on Machine Learning and Systems. pp. 70--79 (2021)

\bibitem{lin2021mcunetv2}
Lin, J., Chen, W.M., Cai, H., Gan, C., Han, S.: Mcunetv2: Memory-efficient patch-based inference for tiny deep learning. arXiv preprint arXiv:2110.15352  (2021)

\bibitem{lin2020mcunet}
Lin, J., Chen, W.M., Lin, Y., Gan, C., Han, S., et~al.: Mcunet: Tiny deep learning on iot devices. Advances in Neural Information Processing Systems  \textbf{33},  11711--11722 (2020)

\bibitem{lin2024awq}
Lin, J., Tang, J., Tang, H., Yang, S., Chen, W.M., Wang, W.C., Xiao, G., Dang, X., Gan, C., Han, S.: Awq: Activation-aware weight quantization for on-device llm compression and acceleration. Proceedings of Machine Learning and Systems  \textbf{6},  87--100 (2024)

\bibitem{lin2015microsoft}
Lin, T.Y., Maire, M., Belongie, S., Bourdev, L., Girshick, R., Hays, J., Perona, P., Ramanan, D., Zitnick, C.L., Dollár, P.: Microsoft coco: Common objects in context (2015)

\bibitem{liu2016ssd}
Liu, W., Anguelov, D., Erhan, D., Szegedy, C., Reed, S., Fu, C.Y., Berg, A.C.: Ssd: Single shot multibox detector. In: Computer Vision--ECCV 2016: 14th European Conference, Amsterdam, The Netherlands, October 11--14, 2016, Proceedings, Part I 14. pp. 21--37. Springer (2016)

\bibitem{liu2015deeplearningfaceattributes}
Liu, Z., Luo, P., Wang, X., Tang, X.: Deep learning face attributes in the wild (2015), \url{https://arxiv.org/abs/1411.7766}

\bibitem{magnitskaya2017electronphononpropertiesnoncentrosymmetric}
Magnitskaya, M., Chtchelkatchev, N., Tsvyashchenko, A., Salamatin, D., Lepeshkin, S., Fomicheva, L., Budzyński, M.: Electron and phonon properties of noncentrosymmetric rhge from ab initio calculations (2017), \url{https://arxiv.org/abs/1708.08788}

\bibitem{sandler2018mobilenetv2}
Sandler, M., Howard, A., Zhu, M., Zhmoginov, A., Chen, L.C.: Mobilenetv2: Inverted residuals and linear bottlenecks. In: Proceedings of the IEEE conference on computer vision and pattern recognition. pp. 4510--4520 (2018)

\bibitem{STM32F}
STMicroelectronics: {STM32F}. \url{https://www.st.com/en/microcontrollers-microprocessors/stm32f412.html} (2024)

\bibitem{STM32H7}
STMicroelectronics: {STM32H7}. \url{https://www.st.com/en/microcontrollers-microprocessors/stm32h7-series.html} (2024)

\bibitem{vadera2022methods}
Vadera, S., Ameen, S.: Methods for pruning deep neural networks. IEEE Access  \textbf{10},  63280--63300 (2022)

\bibitem{xu2020convolutional}
Xu, S., Huang, A., Chen, L., Zhang, B.: Convolutional neural network pruning: A survey. In: 2020 39th Chinese Control Conference (CCC). pp. 7458--7463. IEEE (2020)

\bibitem{yu2016lsunconstructionlargescaleimage}
Yu, F., Seff, A., Zhang, Y., Song, S., Funkhouser, T., Xiao, J.: Lsun: Construction of a large-scale image dataset using deep learning with humans in the loop (2016), \url{https://arxiv.org/abs/1506.03365}

\bibitem{zhou2018semanticunderstandingscenesade20k}
Zhou, B., Zhao, H., Puig, X., Xiao, T., Fidler, S., Barriuso, A., Torralba, A.: Semantic understanding of scenes through the ade20k dataset (2018), \url{https://arxiv.org/abs/1608.05442}

\bibitem{zoph2018learning}
Zoph, B., Vasudevan, V., Shlens, J., Le, Q.V.: Learning transferable architectures for scalable image recognition. In: Proceedings of the IEEE conference on computer vision and pattern recognition. pp. 8697--8710 (2018)

\end{thebibliography}
\end{document}